\documentclass{article}

\usepackage[nonatbib,preprint]{neurips_2021}

\usepackage[utf8]{inputenc} 
\usepackage[T1]{fontenc}    
\usepackage{hyperref}       
\usepackage{url}            
\usepackage{booktabs}       
\usepackage{amsfonts}       
\usepackage{nicefrac}       
\usepackage{microtype}      
\usepackage{xcolor}         
\usepackage{graphicx}
\usepackage{tabularx}
\usepackage{adjustbox}
\usepackage{threeparttable}
\usepackage{subfig}
\usepackage{amsmath}
\DeclareCaptionLabelFormat{singleparen}{#2}
\usepackage[font={small}]{caption}
\usepackage{algorithm}
\usepackage{algorithmic}
\usepackage{multirow}
\usepackage{tabu}
\usepackage{amssymb}

\usepackage{multirow}
\usepackage{tabu}
\usepackage{amssymb}

\newcommand{\red}[1]{{\textcolor{red}{#1}}}

\newcommand{\figref}[1]{Fig.~\ref{#1}}

\newcommand{\tabref}[1]{Table.~\ref{#1}}

\newcommand{\eqnref}[1]{Eq.~(\ref{#1})}
\newcommand{\secref}[1]{Sec.~\ref{#1}}
\renewcommand{\paragraph}[1]{\vspace{1mm}\noindent\textbf{#1}}
\newcommand{\ie}{\textit{i.e.}}
\newcommand{\eg}{\textit{e.g.}}

\makeatletter

\newcommand{\thickhline}{%
    \noalign {\ifnum 0=`}\fi \hrule height 1.5pt
    \futurelet \reserved@a \@xhline
}
\newcolumntype{"}{@{\hskip\tabcolsep\vrule width 1pt\hskip\tabcolsep}}
\makeatother

\hypersetup{
     colorlinks   = true,
     citecolor    = green
}
\hypersetup{linkcolor=red}

\title{Neural Human Performer: Learning Generalizable Radiance Fields for Human Performance Rendering}

\author{Youngjoong Kwon$^{1}$,\ \ \ Dahun Kim$^{2}$,\ \ \ Duygu Ceylan$^{3}$,\ \ \ Henry Fuchs$^{1}$ \\
$^{1}$University of North Carolina at Chapel Hill. \  $^{2}$KAIST. \  $^{3}$Adobe Research. \\
{\tt \small \{youngjoong,fuchs\}@cs.unc.edu\ \ \{mcahny\}@kaist.ac.kr\ \ \{ceylan\}@adobe.com}
}
\begin{document}

\maketitle

\begin{abstract} 
  In this paper, we aim at synthesizing a free-viewpoint video of an arbitrary human performance using sparse multi-view cameras. Recently, several works have addressed this problem by learning person-specific neural radiance fields (NeRF) to capture the appearance of a particular human. In parallel, some work proposed to use pixel-aligned features to generalize radiance fields to arbitrary new scenes and objects. Adopting such generalization approaches to humans, however, is highly challenging due to the heavy occlusions and dynamic articulations of body parts. To tackle this, we propose Neural Human Performer, a novel approach that learns generalizable neural radiance fields based on a parametric human body model for robust performance capture. Specifically, we first introduce a temporal transformer that aggregates tracked visual features based on the skeletal body motion over time. Moreover, a multi-view transformer is proposed to perform cross-attention between the temporally-fused features and the pixel-aligned features at each time step to integrate observations on the fly from multiple views. Experiments on the ZJU-MoCap and AIST datasets show that our method significantly outperforms recent generalizable NeRF methods on unseen identities and poses. The video results and code are available at \href{https://youngjoongunc.github.io/nhp}{https://youngjoongunc.github.io/nhp}.
\end{abstract}

\section{Introduction} \label{introduction}
Free-viewpoint video of a human performer has a variety of applications in the area of telepresence, mixed reality, gaming and etc. Conventional free-viewpoint video systems require extremely expensive setups such as dense camera rigs~\cite{collet2015high, dou2016fusion4d,su2020robustfusion} or accurate depth sensors~\cite{debevec2000acquiring, guo2019relightables}, to capture person-specific appearance information. In this paper, we aim at a scalable solution for free-viewpoint human performance rendering that generalizes across different human performers and requires only sparse camera views. 
However, representing and rendering arbitrary human performances is extremely challenging when the observations are highly sparse (up to three to four views) due to heavy self-occlusions and dynamic articulations of the body parts. In particular, an effective solution needs to coherently aggregate appearance information from sparse multi-view observations across time as the body undergoes a 3D motion. Furthermore, the solution needs to generalize to unseen motions and characters at test time. 

Recently, neural radiance fields (NeRF)~\cite{mildenhall2020nerf,Gao-portraitnerf,li2021neural,park2020deformable,peng2021neural,pumarola2020d,raj2021pva,wang2021ibrnet,xian2020space,yu2020pixelnerf,yuan2020star}
have shown photo-realistic novel view synthesis results in per-scene optimization settings. To avoid the expensive per-scene training and improve the practicality, generalizable NeRFs~\cite{raj2021pva,yu2020pixelnerf,wang2021ibrnet} have been proposed which use image-conditioned, pixel-aligned features and achieve feed-forward view synthesis from sparse input views~\cite{raj2021pva,yu2020pixelnerf}. Direct application of these methods to complex and non-rigid human motion is not straightforward, however, and naive solutions suffer from significant artifacts as shown in \figref{fig:unseen_unseen}. Some existing methods~\cite{saito2019pifu,yu2020pixelnerf} aggregate image features across multiple views by simple average pooling, which often leads to over-smoothed outputs when details observed from multiple views (\eg, front and side views) differ due to self occlusions of humans. Alternatively, several methods~\cite{lombardi2019neural,peng2021neural} have proposed to learn person-specific global appearance features from multi-view observations. However, such methods are not able to generalize to new human performers. 

To address these challenges, we propose Neural Human Performer, a novel approach that learns generalizable radiance fields based on a parametric 3D body model for robust performance capture. In addition to exploiting a parametric body model as a geometric prior, the core of our method is a combination of temporal and multi-view transformers which help to effectively aggregate spatio-temporal observations to robustly compute the density and color of a query point. First, the temporal transformer aggregates trackable visual features based on the input skeletal body motion across time. The following multi-view transformer performs cross-attention between the temporally-augmented skeletal features and the pixel-aligned features for each time step. The proposed modules collectively contribute to the adaptive aggregation of multi-time and multi-view information, resulting in significant improvements in synthesis results in different generalization settings of unseen motions and identities.

We study the efficacy of Neural Human Performer on two multi-view human performance capture datasets, ZJU-MoCap~\cite{peng2021neural} and AIST~\cite{li2021learn}. Experiments show that our method significantly outperforms recent generalizable radience field (NeRF) methods. Furthermore, we compare ours with identity-specific methods~\cite{peng2021neural, thies2019deferred, wu2020multi} that also utilize a 3D human body model prior. Surprisingly, our generalized method achieves better rendering quality than the person-specific dedicated methods when tested on novel poses demonstrating the effectiveness of our transformer-based generalizable representation. 

To summarize, our contributions are:
\begin{itemize}
\item We present a new feed-forward method of synthesizing novel-view videos of arbitrary human performers from sparse camera views. We propose Neural Human Performer that learns generalizable neural radiance representations by leveraging a 3D body motion prior. 
\item We design a combination of temporal and multi-view transformers that can aggregate information on the fly over video frames across multiple views to render each frame of the novel-view video.  
\item We show significant improvements over recent generalizable NeRF methods on unseen identities and poses. Moreover, our generalization results can outperform even person-specific methods when tested on unseen poses.
\end{itemize}

\section{Related works}

\textbf{Human performance capture.} \quad
Novel view synthesis of human performance has a long history in computer vision and graphics. Traditional methods rely on complicated hardware such as dense camera rigs~\cite{collet2015high, dou2016fusion4d,su2020robustfusion} or accurate depth sensors~\cite{debevec2000acquiring, guo2019relightables}. To enable free-view video from sparse camera views, template-based methods~\cite{carranza2003free, de2008performance, gall2009motion, stoll2010video}
exploit pre-scanned human models to track the motion of a person. However, their synthesis results are not photo-realistic and pre-scanned human models are not available in most cases. Recent methods~\cite{natsume2019siclope, saito2019pifu, saito2020pifuhd, zheng2019deephuman} learn 3D human geometry priors along with pixel aligned features to enable detailed 3D human reconstructions even from single images. However, these methods often suffer under complex human poses that are never seen during training and hence cannot be directly used for our purpose of human performance synthesis.

\textbf{Neural 3D representations.} \quad
Recently, there has been great progress in learning neural networks to represent the shape and appearance of scenes. The 3D representations are learned from 2D images via differentiable rendering networks. Convolutional neural networks are used to predict volumetric representations via 3D voxel-grid features~\cite{sitzmann2019deepvoxels, lombardi2019neural, olszewski2019transformable, mescheder2019occupancy, kwon2020rotationally, kwon2020rotationally_temporally}, point clouds~\cite{ aliev2019neural,wu2020multi}, textured meshes~\cite{liao2020towards, liu2019neural, thies2019deferred} and multi-plane images~\cite{ flynn2019deepview,zhou2018stereo}. The learnt representations are projected by a 3D-to-2D operation to synthesize images. However, these methods often have difficulty in scaling to higher resolution due to memory restrictions, and in rendering multi-view consistent images.

To eschew these problems, implicit function-based methods~\cite{liu2020neural,liu2020dist,niemeyer2020differentiable, sitzmann2019scene} learn a multi-layer perceptron that directly translates a 3D positional feature into a pixel generator. The more recent NeRF~\cite{mildenhall2020nerf} learns implicit fields of density and color with a volume rendering technique and achieves photo-realistic view synthesis. Among many following NeRF extensions, ~\cite{park2020deformable, pumarola2020d,yuan2020star,li2021neural} focus on dynamic scenes. While showing impressive results, it's an extremely under-constrained problem to jointly learn NeRF and highly dynamic deformation fields. To regularize the training, Neural Body~\cite{peng2021neural} combines NeRF with a deformable human body model (\eg, SMPL~\cite{loper2015smpl}). 
Despite the promising results, these general NeRF~\cite{li2021neural,yuan2020star} and human-specific NeRF~\cite{Gao-portraitnerf, park2020deformable, peng2021neural, pumarola2020d, xian2020space} methods must be optimized for each new video separately, and generalize poorly on unseen scenarios. Generalizable NeRFs~\cite{raj2021pva, wang2021ibrnet,yu2020pixelnerf} try to avoid the expensive per-scene optimization by image-conditioning using pixel-aligned features. However, directly extending such methods to model complex and dynamic 3D humans is not straightforward when available observations are highly sparse. Unlike existing works, our method exploits temporal and multi-view information on-the-fly and achieves free-viewpoint human rendering in a \textit{feed-forward} manner, also generalizing to new, unseen human identities and poses.

\section{Method}

\begin{figure}[t]
\centering
\def\arraystretch{0.5}
\includegraphics[width=0.94\linewidth]{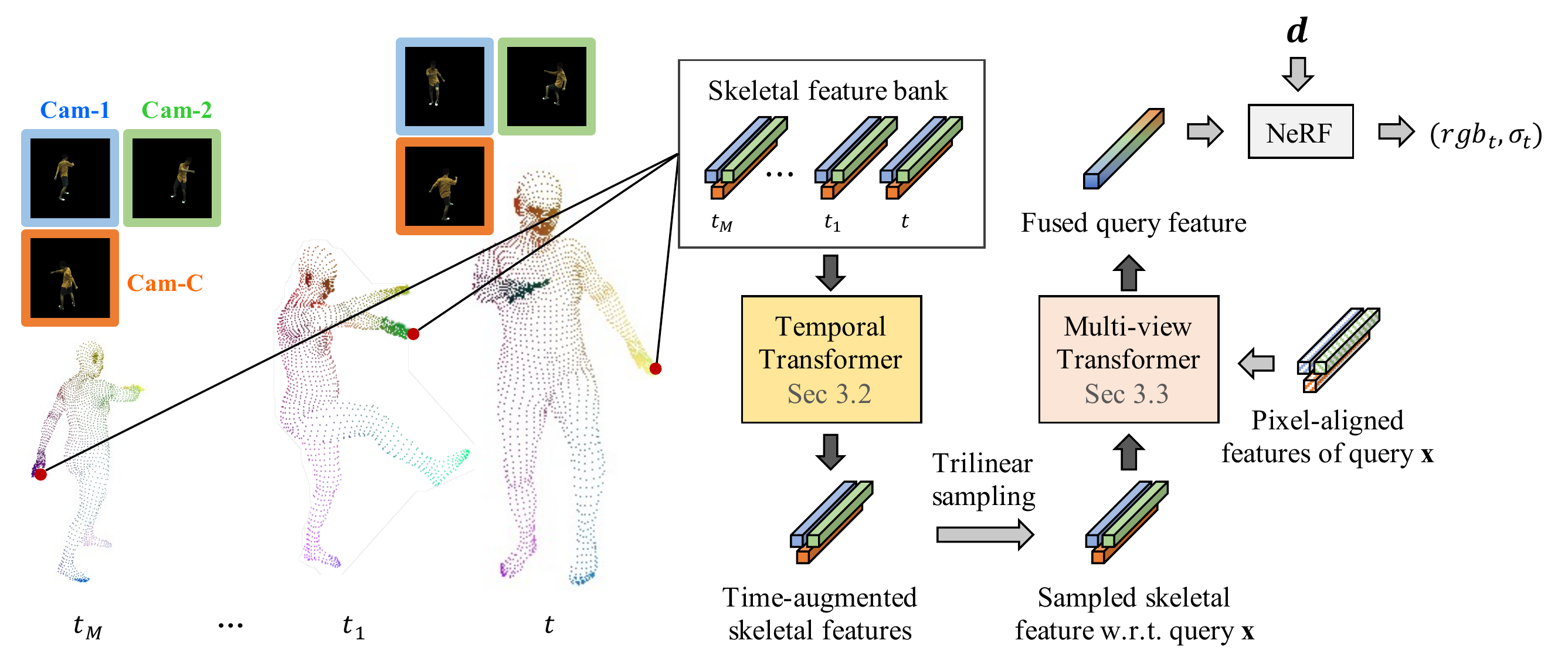}
\caption{\small{\bf Overview of Neural Human Performer.} }
\label{fig:skeletal_feature}
\vspace{-2mm}
\end{figure}

\subsection{Neural Human Performer}
\textbf{Problem definition.} \quad In our setting, given a sparse set (e.g., 3 or 4) of multi-view cameras $c = 1, \dots, C$, input videos of an arbitrary human performance $I_{c,1:T} := \{I_{c,1}, I_{c,2}, ..., I_{c,T}\}$ are captured for each camera view $c$ defined by $\{\mathbf{K}_c, \left[\mathbf{R|t}\right]_c\}$. We assume that the 3D human body model fit corresponding to each frame is given. Our goal is to synthesize a novel view video ${\hat{I}}_{q,1:T}$ for a query viewpoint $q$ defined by $\{\mathbf{K}_q, \left[\mathbf{R|t}\right]_q\}$.

\textbf{Overview.} \quad
To compensate for the sparsity of available input views, we propose to exploit temporal information across video frames. In practice, we sample $M$ memory frames from the original input videos to augment each queried timestep $t$. Our goal is to learn generalizable 3D representations of human performers from multi-time ($M$) and multi-view ($C$) observations.

To this end, the Neural Human Performer is proposed with two main components. The overview is illustrated in \figref{fig:skeletal_feature}. First, we construct the time-augmented skeletal features $\{s\}$. We exploit a human body model (SMPL~\cite{loper2015smpl}) to construct 3D skeletal features by projecting all the SMPL vertices onto each memory frame and picking up the pixel-aligned image features~\cite{raj2021pva, saito2019pifu, yu2020pixelnerf} at the projected 2D locations. The skeletal features are sampled from all memory frames to construct the skeletal feature bank. Inspired by Transformers\cite{carion2020end,vaswani2017attention, wang2018non}, we propose a temporal Transformer that aggregates these memory features into the time-augmented skeletal features $\{s'\}$.

In the second stage, given a query 3D point $\mathbf{x}$, skeletal features are sampled at $\mathbf{x}$. In addition, pixel-aligned features $\{p\}$ at each time $t$ are sampled by directly projecting $\mathbf{x}$ onto the input images $\{I\}_{t}$. The multi-view Transformer is proposed to learn the correlation between the pixel-aligned features $\{p\}$ and the time-augmented skeletal features $\{s'\}$, and to adaptively fuse multi-view information.

 Finally, the output representation of the query point $\mathbf{x}$ is fed into the radiance field module to become the color and density values.
 
\subsection{Construction of time-augmented skeletal features}
\label{section:skeletal} 

For video inputs of moving characters, compared to the static scenes, there are inherently more visual cues as the occluded regions in a frame may be visible in other (potentially distant) frames. To take advantage of the temporal information, we first build up the skeletal feature bank from memory frames by leveraging a parametric body model (see \figref{fig:skeletal_feature}). Then, we propose a temporal Transformer module that aggregates the collected time information.

For each view $c$ and time $t$, we first build frame-level skeletal features $s_{c,t} \in \mathbb{R}^{L\times d}$ by sampling image features at the SMPL vertices' $\mathbb{R}^{3\rightarrow 2}$ projected locations on $I_{c,t} \in \mathbb{R}^{H \times W \times 3}$. $L$ denotes the number of SMPL vertices and $d$ the dimension of image features. 

After collecting all the skeletal features from all memory frames, we propose to aggregate the time information in an attention-aware manner, instead of using simple average pooling.

For any $i^{th}$ skeletal feature vertex $s_{c,t}^i \in \mathbb{R}^{d}$, the proposed temporal Transformer casts attention over all other features contained in the vertex's memory bank $s_{c,t_1:t_M}^i=\{s_{c,t_1}^i, s_{c,t_2}^i, ..., s_{c,t_M}^i\} \in \mathbb{R}^{M\times d}$. In particular, soft weights of all memory feature vertices are computed in a non-local manner with respect to the current timestep $t$. Then, the values of the memory features are weighted summed as
\begin{equation}
\begin{split}
&t\_att_{c,t}^i = \psi \big({\frac{1}{\sqrt{d_0}}} {q(s_{c,t}^i)} \cdot {k{(s_{c,t_1:t_M}^{i})}^{T}} \big),
\quad t\_att_{c,t}^i \in \mathbb{R}^{1\times M} \\ 
&{{s^{\prime}_{c,t}}^i = t\_att_{c,t}^i \cdot v(s_{c,t_1:t_M}^i) + s_{c,t}^i, 
\quad\quad\quad s^{\prime}_{c,t} \in \mathbb{R}^{L\times d}} \quad \forall i
\label{eqn:temporal}
\end{split}
\end{equation}
where $\psi$ represents the softmax operator along the second axis, $q(\cdot)$, $k(\cdot)$ and $v(\cdot)$ are learnable query, key and value embedding functions $\mathbb{R}^{d \rightarrow d_0}$ of the temporal Transformer.

In other words, the representation $s_{c,t}^i$ of each skeletal vertex at time $t$ is computed through a dynamically weighted combination of all its previous and next representations in the memory frames. This allows our network to incorporate helpful information and ignore irrelevant ones from other timesteps. In practice, the temporal Transformer operation in \eqnref{eqn:temporal} is performed by a batch matrix multiplication for all skeletal vertices $L$ and all available viewpoints $C$.

\subsection{Multi-view aggregation of skeletal and query features}

\label{section:pixel}

Given a query 3D point $\mathbf{x} \in \mathbb{R}^3$, we retrieve the corresponding (time-augmented) skeletal feature ${s'_{c,t}}^{\mathbf{x}} \in \mathbb{R}^{d}$ at the queried location via trilinear interpolation in the SMPL space with SparseConvNet~\cite{liu2015sparse}, following  \cite{shi2020pv,peng2020convolutional,yan2018second,peng2021neural}.

In addition, we sample pixel-aligned image feature $p_{c,t}^{\mathbf{x}}$ via direct $\mathbb{R}^{3 \rightarrow2}$ projection of the query point $\mathbf{x}$ on $I_{c,t}$. It is important to note that the pixel-aligned feature $p_{c,t}^{\mathbf{x}}$ is time-specific and represents the exact query location of $\mathbf{x}$, while the skeletal feature ${s'_{c,t}}^{\mathbf{x}}$ is time-augmented (w.r.t $t$) and contains inherent geometric deviations in the SMPL vertices and the following trilinear interpolations. We propose to combine these two complementary features, which will be shown to be effective in~\secref{sec:ablation}.

Given the two sets of multi-view features, skeletal ${s'_{1:C,t}}^\mathbf{x} = \{{s'_{c,t}}^\mathbf{x}|c=1,...,C\} \in \mathbb{R}^{C\times d}$ and pixel-aligned $p_{1:C,t}^\mathbf{x} = \{p_{c,t}^\mathbf{x}|c=1,...,C\} \in \mathbb{R}^{C\times d}$, we propose a multi-view Transformer that performs cross-attention from skeletal to pixel-aligned features. Specifically, the values of pixel-aligned features from all viewpoints is re-weighted based on how much compatible they are with each skeletal features. The non-local cross-attention $mv\_att$ is constructed as:
\begin{equation}
\begin{split}
&mv\_att_{t}^\mathbf{x} = \psi \big({\frac{1}{\sqrt{d_1}}} {k({s'_{1:C,t}}^\mathbf{x})} \cdot {k(p_{1:C,t}^\mathbf{x})^T} \big),
\quad mv\_att_{t}^\mathbf{x} \in \mathbb{R}^{C\times C} \\ 
&{z_{1:C,t}^\mathbf{x} = mv\_att_{t}^\mathbf{x} \cdot v(p_{1:C,t}^\mathbf{x}) + v({s'_{1:C,t}}^\mathbf{x}), 
\quad\quad\quad z_{1:C,t}^\mathbf{x} \in \mathbb{R}^{C\times d}, \quad z_{c,t}^\mathbf{x} \in \mathbb{R}^{1\times d}}
\label{eqn:multiview}
\end{split}
\end{equation}

where $\psi$ represents the softmax operator along the second axis.
Note that $k$ and $v$ are new layers independent from those in the temporal Transformer.
The confident observations in each view will have large weights and be highlighted, and vice versa. Finally, we use the view-wise mean of 
$z_{t}^\mathbf{x} = \frac{1}{C}\sum_c {z_{c,t}^\mathbf{x}} \in \mathbb{R}^{d}$ as our \textit{meta-time} and \textit{meta-view} representation of the query point $x$.

The final density $\sigma_t(\mathbf{x})$ and color values ${rgb}_t(\mathbf{x})$ at time $t$ are computed as:
\begin{equation}
\begin{split}
\sigma_t(\mathbf{x}) = MLP_{\sigma}(z_t^{\mathbf{x}}), \quad \quad
{rgb}_t(\mathbf{X}) = MLP_{\mathbf{rgb}}(\sum_c{(z_{c,t}^{\mathbf{x}}; \gamma_\mathbf{d}(\mathbf{d}))}/C),
\end{split}
\end{equation}
where $MLP_{\sigma}$ and $MLP_{\mathbf{rgb}}$ consist of four and two linear layers respectively, and $\gamma_\mathbf{d} : \mathbb{R}^{3 \rightarrow {6\times l}}$ is a positional encoding of viewing direction $\mathbf{d} \in \mathbb{R}^3$ as in \cite{mildenhall2020nerf} with $2 \times l$ different basis functions. 

More details on the network architecture can be found in the supplementary material.

\subsection{Volume Rendering}
\label{section:render}
The predicted color of a pixel $p \in \mathbb{R}^2$ for a target viewpoint $q$ in the focal plane of the camera and center $\mathbf{r}_0 \in \mathbb{R}^3$ is obtained by marching rays into the scene using the camera-to-world projection matrix, $
\mathbf{P}^{-1} = [\mathbf{R}_q|\mathbf{t}_q]^{-1}\mathbf{K}_q^{-1} 
$ with the direction of the rays given by $\mathbf{d} = \frac{\mathbf{P}^{-1}p - \mathbf{r}_0 }{\lVert{\mathbf{P}^{-1}p - \mathbf{r}_0}\rVert}.$

We then accumulate the radiance and opacity along the ray $\mathbf{r}(z) =\mathbf{r}_0 + z\mathbf{d}$ for $z \in [z_{\mathrm{near}},z_{\mathrm{far}}]$ as defined in NeRF~\cite{mildenhall2020nerf} as follows:
\begin{equation}
    \mathbf{I}_{q}(p) = \int_{z_{\mathrm{near}}}^{z_{\mathrm{far}}} \mathbf{T}(z)\sigma(\mathbf{{r}}(z))\mathbf{c(r}(z),\mathbf{d}) dz, \quad where\quad 
    \mathbf{T}(z) = \operatorname{exp}\left(-\int_{z_{\mathrm{near}}}^{z} \mathbf{\sigma(r}(s))ds\right)
\end{equation}

In practice, we uniformly sample a set of $64$ points $z\sim[z_{near}, z_{far}]$. We set $\mathbf{X}=\mathbf{r}(z)$ and use the quadrature rule to approximate the integral. We compute the 3D bounding box of the SMPL parameters at time $t$ and derive the bounds for ray sampling $z_{near}, z_{far}$.

\subsection{Loss Function}
For ground truth target image $\mathbf{I}_{q,t}$, we train both the radiance field and feature extraction network using a simple photo-metric reconstruction loss $\mathcal{L} = \lVert\mathbf{\hat{I}}_{q,t} - \mathbf{I}_{q,t} \rVert_2 ~.$

\section{Experiments}

We present novel view synthesis and 3d reconstruction results of human performances in different generalization scenarios. We compare our method against the current best view-synthesis methods from two classes: body model-based, per-scene optimization methods (\secref{sec:per_scene}) and generalizable NeRF methods (\secref{sec:generalizable}). 
We experiment on ZJU-MoCap~\cite{peng2021neural} and AIST datasets~\cite{tsuchida2019aist, li2021learn}. For training and testing of our model as well as the baselines, we remove the background using the foreground mask that is either provided by the dataset or pre-computed.
Unless otherwise specified, we sample two memory frames $\{t-20, t+20\}$ at time $t$ (total three timesteps) and take three canonical input views in all experiments. 
The details of the datasets, training process, additional results and video results are provided in the supplementary material.

\subsection{Comparison with body model-based, per-scene optimization methods.}
\label{sec:per_scene}
\textbf{Baselines.} \quad For body model-based methods, we compare with the state-of-the-art Neural Body (NB) \cite{peng2021neural} that combines SMPL and NeRF in a per-scene optimization setting. Neural Textures (NT) \cite{thies2019deferred} renders a coarse mesh with latent texture maps and uses a 2D CNN to render target images. We use the SMPL mesh as the input mesh. NHR \cite{wu2020multi} extracts 3D features from input point clouds and renders them into 2D images. Since dense point clouds are difficult to obtain from sparse camera views, we take SMPL vertices as input point clouds. These methods have reported that their learnt per-model representations can adapt to new poses of the same performer, \ie, novel pose synthesis.

\begin{table}[]
 \centering
  \subfloat[. Test results on \textbf{source} models' \\ \textbf{unseen} poses]{%
  \resizebox{.32\textwidth}{!}{
    \begin{tabular}{l rrrr}
    \hline
        {Method}        &&&  PSNR   & SSIM    \\
        \thickhline
         \multicolumn{5}{l}{Trained on \textbf{source} models} \\ \hline
        NB          &&& 23.79  & 0.887   \\
        NHR         &&& 22.31  & 0.871   \\
        NT          &&& 22.28  & 0.872   \\
        \thickhline
        \multicolumn{5}{l}{Trained on \textbf{source} models} \\ \hline
        Ours        &&& \textbf{26.94}  & \textbf{0.929}   \\
        \hline
        
    \end{tabular}
    \label{table:pose}
    }
  }
  \subfloat[. Test results on \textbf{target} models' \\ \textbf{unseen} poses]{%
  \resizebox{.32\textwidth}{!}{
    \begin{tabular}{l rrrr}
        \hline
        {Method}      &&&  PSNR   & SSIM   \\
        \thickhline
        \multicolumn{5}{l}{Trained on \textbf{target} models} \\ \hline
        NB            &&& 22.88  & 0.880   \\
        NHR           &&& 22.03  & 0.875   \\
        NT            &&& 21.92  & 0.873   \\
        \thickhline
        \multicolumn{5}{l}{Trained on { \textbf{source}} models} \\ \hline
        Ours          &&& \textbf{24.75}  & \textbf{0.906}   \\
        \hline
        
    \end{tabular}
    \label{table:identity}
    }
  }
  \subfloat[. Test results on \textbf{source} models' \\ \textbf{seen} poses]{%
  \resizebox{.35\textwidth}{!}{
    \begin{tabular}{l rrrr}
    \hline
        {Method}      &&&  PSNR   & SSIM    \\
        \thickhline
        \multicolumn{5}{l}{Trained on \textbf{source} models} \\ \hline
        NB            &&& 28.51  & \textbf{0.947}   \\
        NHR           &&& 23.95  & 0.897   \\
        NT            &&& 23.86  & 0.896   \\
        \thickhline
        \multicolumn{5}{l}{Trained on \textbf{source} models} \\ \hline
        Ours          &&& \textbf{28.73}  & {0.936}   \\
        Ours          &\multicolumn{2}{l}{\textit{per-scene}}& \textbf{31.57}  & {0.966}   \\
        \hline
        
    \end{tabular}
    \label{table:seen}
    }
  }
\vspace{2mm}
\caption{\small{\textbf{Comparison with other body model-based, per-scene optimization methods.}}}
\label{table:body}
\vspace{-2mm}
\end{table}

\begin{figure}[t]
\centering
\def\arraystretch{0.5}
\includegraphics[width=1.0\linewidth]{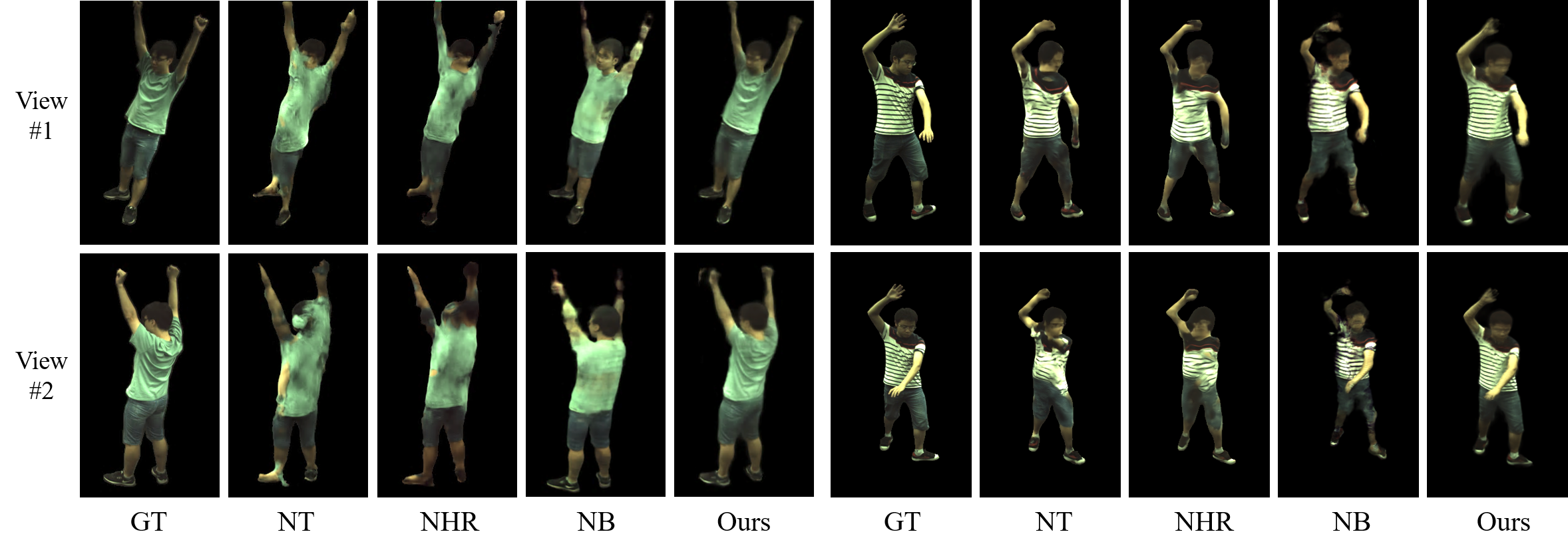} \\
\vspace{-2mm}
\caption{\small{\bf Pose generalization -- comparison with other body model-based, per-person optimization methods.} Results of NT: Neural textures~\cite{thies2019deferred}, NHR: Neural human rendering~\cite{wu2020multi}, NB: Neural body~\cite{peng2021neural} and ours. Novel view synthesis on ZJU-MoCap. Tested on \textbf{source} models' \textbf{unseen} poses (All methods are trained on \textbf{source} models; Competing methods are trained in a per-person manner.)}
\label{fig:seen_unseen}
\vspace{-3mm}
\end{figure}

\textbf{Setup.} \quad We experiment with ZJU-MoCap dataset \cite{peng2021neural} which provides performance captures of 10 human subjects captured from 23 synchronized cameras, human body model parameters as well as the foreground mask corresponding to each frame. Each video contains complicated motions such as kicking and Taichi and a length between 1000 to 2000 frames. 
We consider three different comparison settings as detailed below. We first split the dataset into two parts: source and target videos. In all comparisons, the first 300 frames of either source or target videos are used during training, and the remaining next frames (unseen poses) are held out for testing. Note that the baseline methods are always trained in a per-model manner. To validate whether the training is reproducible, we experiment with 5 independent runs with random train/test splits and observe a variance of 0.15 PSNR, showing that the results are quite robust. In each independent run, we used seven models for training and the other three for testing.

\begin{figure}[t]
\centering
\def\arraystretch{0.5}
\includegraphics[width=1.0\linewidth]{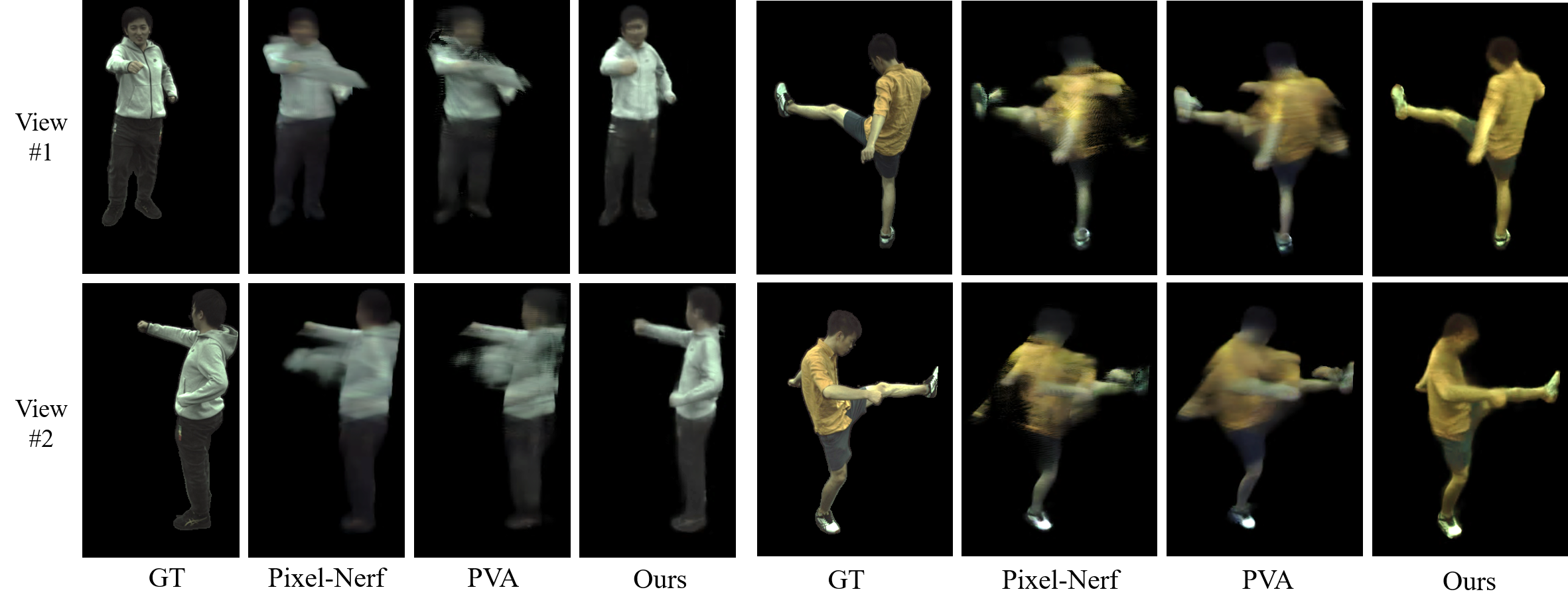}
\vspace{-6mm}
\caption{\small{\bf Identity-and-pose generalization -- comparison with generalizable NeRF methods.} Results of Pixel-Nerf~\cite{yu2020pixelnerf}, PVA: Pixel-aligned volumetric avatar~\cite{raj2021pva} and ours. Novel view synthesis on ZJU-MoCap. Tested on \textbf{target} models' \textbf{unseen} poses (All methods are trained on \textbf{source} models.)}
\label{fig:unseen_unseen}
\vspace{-4mm}
\end{figure}

\begin{table}[t]
 \centering
  \subfloat[. Generalization results on ZJU-MoCap.]{%
  \resizebox{.38\textwidth}{!}{
    \begin{tabular}{ll|c|c}
        \hline
        Method       &&  PSNR   & SSIM    \\
        \thickhline
        Pixel-NeRF   && 23.17  & 0.8693   \\
        PVA          && 23.15  & 0.8663   \\
        \hline
        Ours         && \textbf{24.75}  & \textbf{0.9058}  \\
        \hline
        
    \end{tabular}
    \label{table:ZJU}
    }
  } \hspace{8mm}
  \subfloat[. Generalization results on AIST.]{%
  \resizebox{.38\textwidth}{!}{
    \begin{tabular}{ll|c|c}
        \hline
        Method       &&  PSNR   & SSIM     \\
        \thickhline
        Pixel-NeRF   && 18.06  & 0.7304   \\
        PVA          && 17.82  & 0.7211   \\
        \hline
        Ours         && \textbf{19.03}  & \textbf{0.8390}  \\
        \hline
        
    \end{tabular}
    \label{table:AIST}
    }
  }
\vspace{2mm}
\caption{\small{\textbf{Comparison with generalizable NeRF methods.}}}
\label{table:general}
\vspace{-7mm}
\end{table}

\textbf{Results.} \quad We present three different comparison settings to validate our method. We would like to point out that all the comparison settings place our method (`ours') in disadvantage. This is because our model is trained on all the source subjects at once (one network for all subjects), while the competing methods are per-subject trained on the \textit{subject to be tested} (one network for one subject) - easier setting. First, we evaluate \textbf{1) Pose generalization} (\tabref{table:pose} and \figref{fig:seen_unseen}). For all methods, we train on source models, and test on the same source models' unseen poses. Our method significantly outperforms all the baselines and the state-of-the-art Neural Body~\cite{peng2021neural} by \textbf{+3.15} PSNR and +0.042 SSIM scores. We also present a very challenging setting: \textbf{2) Identity generalization} (\tabref{table:identity}). Our method is trained on source models, while other baselines are trained on target models. Then, all methods are tested on the target models' unseen poses. Note that this comparison is \textit{disadvantageous to ours} since the competing methods have \textit{seen} the testing models as they must be trained separately per human model (no identity generalization for baselines). Surprisingly, our unseen-model generalization outperforms all per-scene optimized baselines by a health margin of +1.87 PSNR and +0.026 SSIM scores. 
The comparison results 1 and 2 indicate that our improvements are not merely from the use of body model prior (SMPL), but that our proposed architecture with the temporal as well as the multi-view transformer can generalize well onto the novel identities and poses, and can produce photo-realistic results. 
In \tabref{table:seen}, we show the \textbf{3) performance on seen model with seen pose}, where all the methods are trained on the source models and tested on the seen trained poses. Our method (`ours') shows comparable results with the state-of-the-art per-scene optimization method~\cite{peng2021neural}.
Finally, to make a more fair comparison, we conduct an additional experiment in the exact same per-scene training and testing setting as the other competing methods. \tabref{table:seen} shows our model (`ours per-scene') achieves significant improvements over the best-performing baseline~\cite{peng2021neural} by +3 PSNR and +1.4\% SSIM. 

\subsection{Comparison with generalizable NeRF methods.}
\label{sec:generalizable}
\textbf{Baselines.} \quad Among the recent generalizable NeRF methods~\cite{raj2021pva, yu2020pixelnerf, wang2021ibrnet}, we compare with Pixel-NeRF~\cite{yu2020pixelnerf} and PVA~\cite{raj2021pva} which focus on very sparse (up to 3 or 4) input views. 
we reimplement \cite{raj2021pva} since it is not open-sourced.

\textbf{Setup.} \quad In addition to ZJU-MoCap (details are in \secref{sec:per_scene}), we experiment on larger AIST dataset \cite{tsuchida2019aist,li2021learn} to further evaluate different methods' generalization abilities. AIST dataset provides dance videos of 30 human subjects captured from 9 cameras, together with the SMPL parameters. We extract the foregroud mask of each image using an off-the-shelf human parser \cite{gong2018instance}. AIST dataset contains highly diverse motions, slow to fast, simple to complicated.
We split the dataset into 20 and 10 subjects for training and testing respectively, where the testing dataset contains novel models and novel poses.

\begin{figure}[t]
\centering
\def\arraystretch{0.5}
\includegraphics[width=1.0\linewidth]{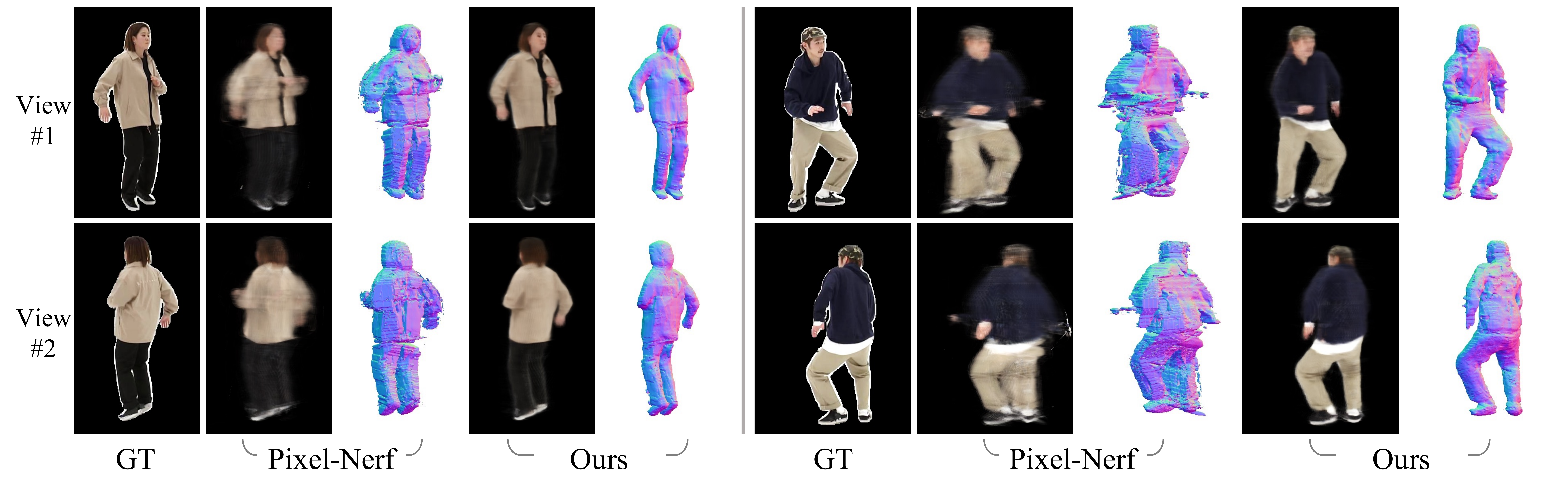} \\
\vspace{-3mm}
\caption{{\small{\bf Generalization results on AIST.} Novel view synthesis and 3D reconstruction results on unseen models' unseen poses.}}
\label{fig:aist}
\vspace{-4mm}
\end{figure}

\textbf{Novel view synthesis results.} \quad   \tabref{table:general} shows the comparison. For all datasets and all metrics, our method consistently outperforms the baselines by healthy margins of +1.6 PSNR and +0.037 SSIM scores. \figref{fig:unseen_unseen} and \figref{fig:aist} present the same tendency in visualizations. Pixel-NeRF and PVA aggregate multi-view observations via average pooling without explicitly considering the correlation between the views. In contrast, our temporal and multi-view transformers learn to model the correlation between input views and integrate different observations to help the NeRF module to produce more accurate results. Another advantage of our method is that the used body model prior provides a robust geometric cue to handle the self-occlusion of human subjects.

\begin{figure}[t]
\centering
\def\arraystretch{1.0}
\includegraphics[width=1.0\linewidth]{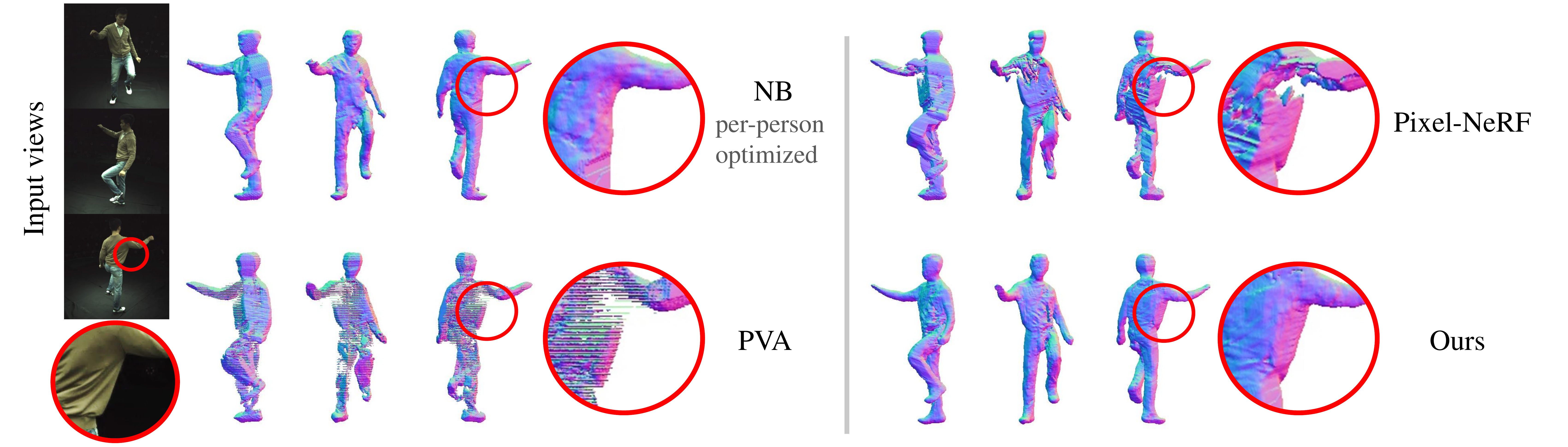}
\caption{{\small{\bf 3D reconstruction on ZJU-MoCap.} Tested on unseen model's unseen pose except Neural Body (per-person optimized). NB: Neural Body~\cite{peng2021neural}, PVA: Pixel volumetric avatar~\cite{raj2021pva}, Pixel-NeRF~\cite{yu2020pixelnerf} and ours.}}
\label{fig:mesh_viz}
\vspace{-1mm}
\end{figure}

\textbf{3D reconstruction results.} \quad We also evaluate 3D reconstruction of generalizable NeRF methods and our method on ZJU-MoCap (\figref{fig:mesh_viz}) and AIST datasets (\figref{fig:aist}) given three input views. The visualization shows that our 3D reconstruction aligns well with the input image, and is more reliable than even the per-person method~\cite{peng2021neural} (\eg, the shape of upper cloth in ~\figref{fig:mesh_viz}).

Overall, these results indicate that as human models are complex and occlusion-heavy, more sophisticated designs than simple image-conditioning are required to learn robust and accurate 3D human representations.

\subsection{Cross-dataset generalization.}
\label{sec:cross-dataset}
We further further study the generalization ability of our method across different datasets.
\tabref{table:cross_dataset} shows the cross-dataset generalization results. We would like to first point out that the two datasets~\cite{li2021neural, tsuchida2019aist} have highly different color distribution (background, lighting) and distance of camera to the subject, making the cross-dataset generalization extremely challenging.
Nevertheless, we found that only 8-minute fine-tuning on the target dataset can already outperform the baselines fully-trained on the target dataset, and 16-minute fine-tuning performs on par with our model fully-trained on the target dataset.

\subsection{Ablation studies}
\label{sec:ablation}
\tabref{table:ablation} shows the ablation study on ZJU-MoCap on unseen identities and unseen poses, using three time-steps and three camera views as input. Note that all the items without either temporal or multi-view transformer module use simple average pooling instead, to fuse temporal or multi-view observations respectively.

\textbf{Complementariness of skeletal and pixel-aligned query features.} \quad
`Sk' uses only time-augmented skeletal features (\secref{section:skeletal}) without time-specific pixel-aligned features, while `Px' uses only the time-specific pixel-aligned features, on the contrary. Both `only' models show the largest drops compared to our full model, and `Sk + Px' model improves them by +1.2 PSNR and +0.9 PSNR respectively. This validates the complementariness between the skeletal and pixel-aligned features in that one is time-augmented but involves slight geometric deviations, while another is time-specific and represents exact query location, as discussed in ~\secref{section:pixel}.

\textbf{Impact of temporal and multi-view transformers.} \quad 
`Sk + Px' uses no transformers so far, falling behind our full model by -1.3 PSNR score. Then `Sk + Px + T' adds the temporal transformer and improves +0.7 PSNR score, showing its effectiveness in aggregating information over video frames. `Sk + Px + MV' uses the multi-view transformer module and shows the largest gain of +1.0 PSNR, indicating the efficacy of learnt cross-attention between the skeletal features and pixel-aligned features, as well as the importance of learnt inter-view correlations. Our full model `Sk + Px + T + MV' shows the best use of all the proposed components and achieves 24.75 PSNR and 0.9058 SSIM.

\begin{table}[t]
 \centering
  \resizebox{\textwidth}{!}{%
    \begin{tabular}[]{l |c c | c c | c c}
        \hline
        Variant                & Skeletal      &   Pixel-aligned &   T-transformer    &   MV-transformer    & PSNR  & SSIM \\
        \thickhline
          Sk            &   \checkmark  &               &                  &                   & 22.31 & 0.8865 \\
          Px            &               &  \checkmark      &               &                   & 22.58 & 0.8780 \\ 
        \hline
        
          Sk + Px             &  \checkmark   & \checkmark    &               &                & 23.47 & 0.8906 \\ 
        
          Sk + Px + T         &  \checkmark   & \checkmark    &   \checkmark  &                & 24.21 & 0.9016 \\
        
         Sk + Px + MV         &  \checkmark   & \checkmark    &               &   \checkmark   & 24.44 & 0.9034 \\
        \hline
        
          Sk + Px + T + MV    &  \checkmark   & \checkmark    &  \checkmark   &   \checkmark   & \textbf{24.75} & \textbf{0.9058} \\
        \hline
    \end{tabular}
  }
\caption{\small{\textbf{Ablation study.} Generalization results on ZJU-MoCap. Sk: skeletal features, Px: pixel-aligned features, T: temporal transformer, MV: multi-view transformer. }}
\label{table:ablation}
\vspace{-5mm}
\end{table}

\begin{table}[t]
 \centering
  \subfloat[. Cross-dataset generalization.]{%
  \resizebox{.54\textwidth}{!}{
    \begin{tabular}{l|c|c|c}
    \hline
    Exp. protocol           & Fine-tune & PSNR  & SSIM   \\ \thickhline
    Trained on AIST         & 8-min     & 24.25 & 0.8954 \\ \cline{2-4} 
    Fine-tuned on ZJU-Mocap & 16-min    & 24.73 & 0.9023 \\ \hline
    Trained on ZJU-Mocap    & 8-min     & 18.63 & 0.8242 \\ \cline{2-4} 
    Fine-tuned on AIST      & 16-min    & 18.83 & 0.8374 \\ \hline

    \end{tabular}
    \label{table:cross_dataset}
    }
  } 
  \hspace{3mm}
  \subfloat[. Impact of number of camera views.]{%
  \resizebox{.40\textwidth}{!}{
    \begin{tabular}{c|c|c|c}
        \hline
       \# Timesteps &  \# Views     &  PSNR   & SSIM     \\
        \thickhline
        \multirow{4}{*}{1} & 1            & 20.13  & 0.835   \\
                            &2            & 21.82  & 0.871   \\
                            &3            & 23.33  & 0.906   \\
                            &4            & 23.51  & 0.913   \\
        \hline
        
    \end{tabular}
    \label{table:views}
    }
  }
\vspace{1mm}
\caption{\small{Cross-dataset transferability (left) and impact of different number of camera views tested on the ZJU-Mocap unseen subject in unseen poses (right).}}
\label{table:ablation_2}
\vspace{-3mm}
\end{table}

\textbf{Impact of number of camera views.} \quad 
\tabref{table:views} shows our model's testing results with the different number of input views, fixing the number of timesteps as one. Our method degrades reasonably as the input views become very sparse (as few as one).

\section{Limitations} \label{Limitations}

We tackle some shortcomings of existing body model-based and generalizable NeRF methods with a focus on generalizable human performance rendering, but there are challenges yet to be explored. 1) While we show that our method can generalize across datasets with finetuning, the generalization capability will be limited as the distribution of the datasets become significantly different. 2) The performance of our method will be affected as the SMPL parameter accuracy degrades. This is the reason why the scores on the AIST dataset, where the motions are complicated and thus relatively difficult to obtain accurate SMPL parameters, are lower compared to the scores experimented on the ZJU-Mocap dataset. It would be an interesting direction to jointly refine the SMPL parameters using differentiable rendering for in-the-wild applications. 3) Our algorithm does not have an explicit assumption of static cameras. However, in practice, it might be hard to estimate the inputs to our method (SMPL fits, camera parameters, foreground masks) with moving cameras due to motion blur, changing background, lighting, and synchronization issues etc. We consider this as an orthogonal problem and expect that any advancements in unconstrained multi-view capture setups will help to generalize our method to in-the-wild settings.

\section{Societal impact} \label{Societal}
We discuss the potential societal impact of our work. The positive side is that the human performance synthesis is the key component of realizing telepresence, which has become more important especially in this pandemic era. In the future, people physically apart can feel like they are in the same space and feel connected with a few inexpensive webcams and AR/VR headsets thanks to the development of the telepresence. The negative aspect is that it can help organizations easier to identify people by reconstructing them from an only small number of surveillance cameras. We strongly hope that our research could be used in positive directions.

\section{Conclusion}
We present Neural Human Performer, a generalizable radiance field network based on a parametric body model that can synthesize free-viewpoint videos for arbitrary human performers from sparse camera views. Leveraging the trackable visual features from the input body motion prior, we propose a combination of a temporal Transformer and a multi-view Transformer that integrates multi-time and multi-view observations in a feed-forward manner. Our method can produce photo-realistic view synthesis of new unseen poses and identities at test time. In various generalization settings on ZJU-MoCap and AIST datasets, our method achieves state-of-the-art performance outperforming the body model-based per-scene optimization methods as well as the generalizable NeRF methods.

\begin{ack}
We thank Sida Peng of Zhejiang University, Hangzhou, China, for many very helpful discussions on a variety of implementation details of Neural Body~\cite{peng2021neural}. We also thank Ruilong li and Alex Yu of UC Berkeley for many discussions on the AIST++ dataset~\cite{li2021learn} and pixelNeRF~\cite{yu2020pixelnerf} details.
\end{ack}

{\small
\bibliographystyle{plain}
\bibliography{main}
}
\clearpage

\begin{supp}

\end{supp}

\appendix

\section{Video results}
Video results on free-viewpoint rendering and 3D reconstruction with ZJU-MoCap and AIST datasets can be found at \href{https://youngjoongunc.github.io/nhp}{https://youngjoongunc.github.io/nhp}.

\section{Reproducibility}

\subsection{Implementation details.} We describe the implementation details in the interest of reproducibility. Note that due to the high computing cost, we did not spend significant effort to tune the architecture or training procedure, and it is possible that variations can perform better, or that smaller models may suffice. Code will be made public upon publication.

\paragraph{Image feature extractor.}
As briefly discussed in the main paper, we use an ImageNet-pretrained ResNet18 backbone to extract a feature pyramid. For an image of shape H$\times$W, we take the multi-scale feature maps of shapes 

\quad $\{$ 64 $\times$ H/2 $\times$ W/2, \quad 64 $\times$ H/4 $\times$ W/4, \quad 128 $\times$ H/8 $\times$ W/8,
\quad 256 $\times$ H/16 $\times$ W/16 $\}$.

These feature maps are bilinearly upsampled to the highest resolution \ie, H/2 $\times$ W/2, and concatenated into a shape 512 $\times$ H/2 $\times$ W/2.

\paragraph{Temporal transformer.}

\begin{figure}[t]
\centering
\def\arraystretch{0.5}
\includegraphics[width=0.94\linewidth]{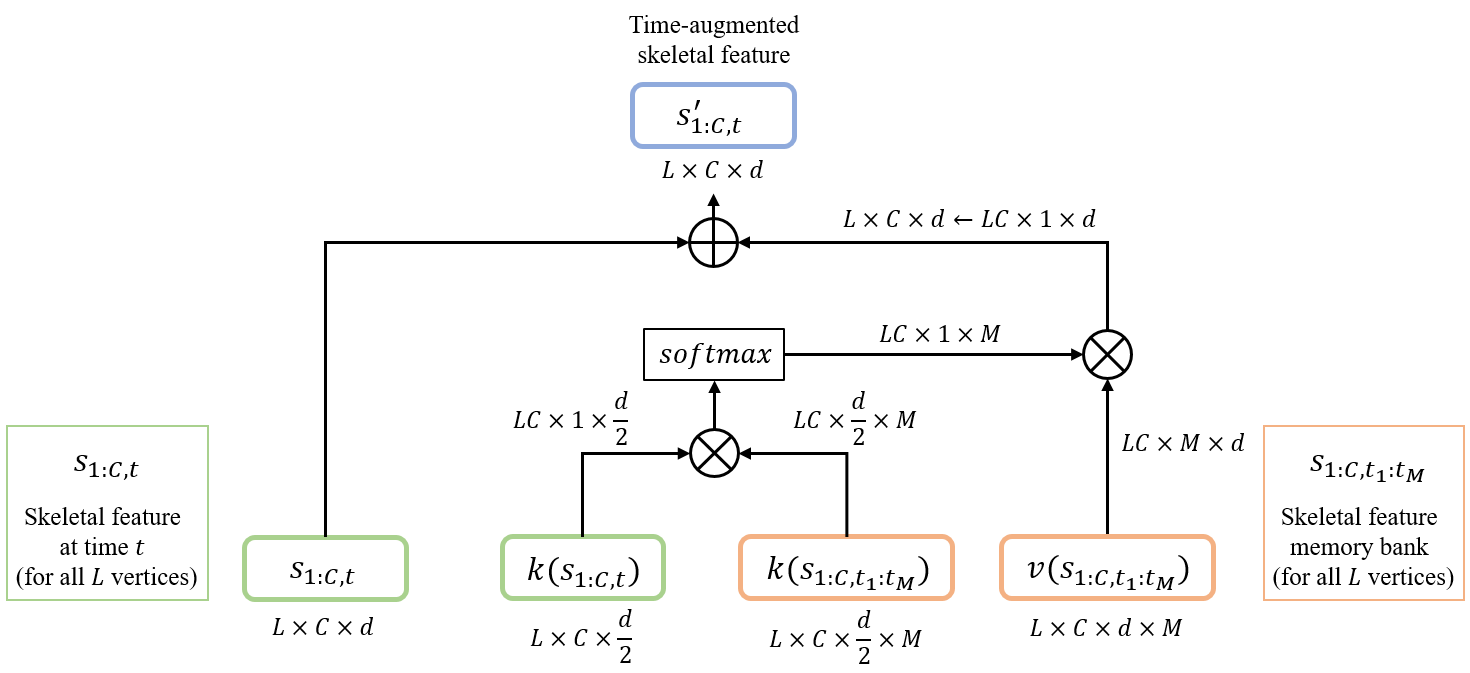}
\caption{\small{\bf Overview of temporal Transformer's attention between the skeletal features at $t$ and skeletal \textit{memory} features. } }
\label{fig:temporal_transformer}
\vspace{2mm}
\end{figure}

The temporal Transformer is used in construction of time-augmented skeletal features in Section \red{3.2} of the manuscript. The overview of the attention between the skeletal feature at $t$ and skeletal \textit{memory} features is illustrated in \figref{fig:temporal_transformer}. All the $L$ vertices are processed batch-wise, where the attention is computed for each vertex. $d$ is set to 64.

\begin{figure}[t]
\centering
\def\arraystretch{0.5}
\includegraphics[width=0.94\linewidth]{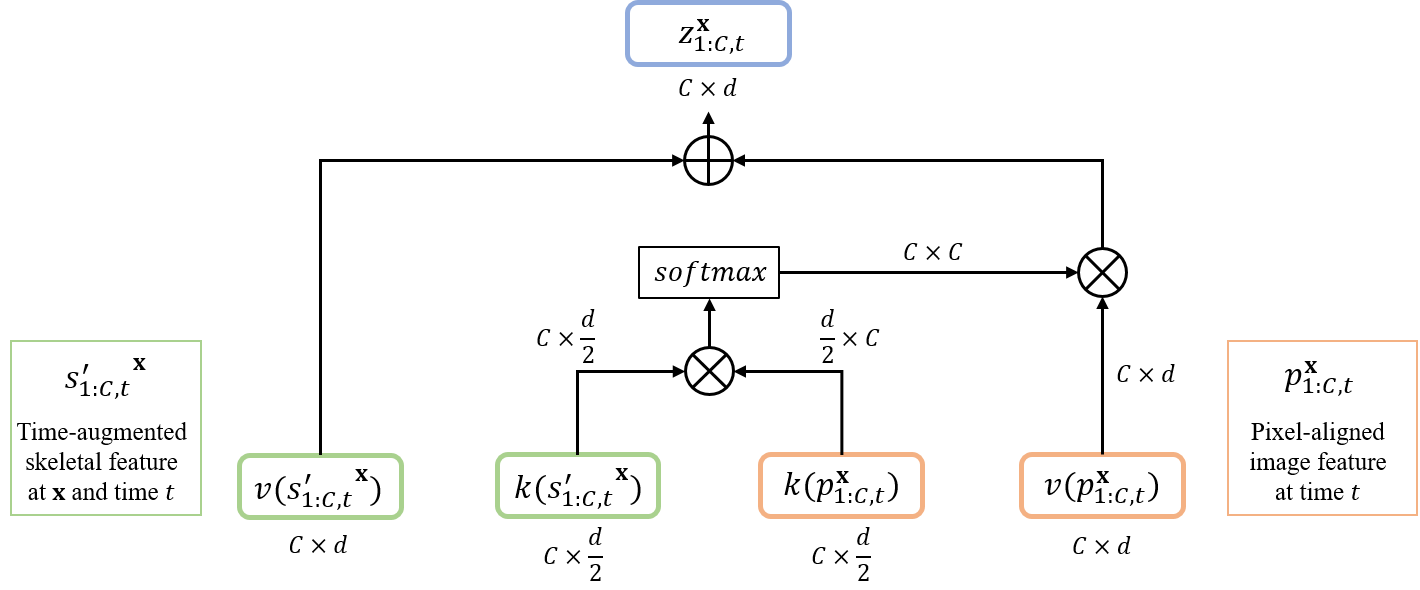}
\caption{\small{\bf Overview of multi-view Transformer's cross-attention between the sampled time-augmented skeletal features and time-specific pixel-aligned feature at $t$.} }
\label{fig:multiview_transformer}
\vspace{-2mm}
\end{figure}

\begin{table}[t]
\centering
\scalebox{0.9}{
\begin{tabular}{l|l|l}
\hline
& Layer Description & Output Dim.  \\ \hline \hline
& Input volume & D' $\times$ H' $\times$ W' $\times$ 64 \\ \hline
1-2 & ($3 \times 3 \times 3$ conv, 64 features, stride 1) $\times$ 2 
&  D' $\times$ H' $\times$ W' $\times$ 64 \\
3 & ($3 \times 3 \times 3$ conv, 64 features, stride 2) 
&  D'/2 $\times$ H'/2 $\times$ W'/2 $\times$ 64 \\
4-5 & ($3 \times 3 \times 3$ conv, 64 features, stride 1) $\times$ 2
&  D'/2 $\times$ H'/2 $\times$ W'/2 $\times$ 64 \\
6 & ($3 \times 3 \times 3$ conv, 64 features, stride 2) 
&  D'/4 $\times$ H'/4 $\times$ W'/4 $\times$ 64 \\
7-9 & ($3 \times 3 \times 3$ conv, 64 features, stride 1) $\times$ 3 
&  D'/4 $\times$ H'/4 $\times$ W'/4 $\times$ 128 \\
10 & ($3 \times 3 \times 3$ conv, 128 features, stride 2) 
&  D'/8 $\times$ H'/8 $\times$ W'/8 $\times$ 128 \\
11-13 & ($3 \times 3 \times 3$ conv, 128 features, stride 1) $\times$ 3 
&  D'/8 $\times$ H'/8 $\times$ W'/8 $\times$ 128 \\
14 & ($3 \times 3 \times 3$ conv, 128 features, stride 2) 
&  D'/16 $\times$ H'/16 $\times$ W'/16 $\times$ 128 \\
15-17 & ($3 \times 3 \times 3$ conv, 128 features, stride 1) $\times$ 3
&  D'/16 $\times$ H'/16 $\times$ W'/16 $\times$ 128 \\
\hline
& Resize \& Concat. outputs of layer 5, 9, 13, and 17 
&  D'/16 $\times$ H'/16 $\times$ W'/16 $\times$ 384 \\

\hline

\end{tabular}
} 

\vspace{3mm}

\caption{\textbf{Architecture of SparseConvNet.} Each layer consists of sparse convolution, batch normalization and ReLU.}
\label{tab:sparse_conv}
\end{table}

\paragraph{Sampling of time-augmented skeletal feature w.r.t. a query point $\mathbf{x}$.}

When we are given a query point $\mathbf{x}$ in 3D space, we sample the corresponding feature at $\mathbf{x}$'s 3D location, ${s'_{1:C,t}}^{\mathbf{x}} \in \mathbb{R}^{C\times d}$,  from the previously constructed time-augmented skeletal features $s'_{1:C,t} \in \mathbb{R}^{L\times C\times d}$. Inspired by \cite{yan2018second, shi2020pv, peng2020convolutional,peng2021neural}, we adopt the SparseConvNet \cite{liu2015sparse} to perform such sampling, whose architecture is described in Table~\ref{tab:sparse_conv}. First, we compute the 3D bounding box of the human body based on the SMPL parameters, and divide the 3D box into small voxels of size of $5mm \times 5mm \times 5mm$, resulting in a $D'\times H'\times W'$ (depth, height, width) volume. The SparseConvNet consists in 3D sparse convolutions to process the input volume, diffusing the skeletal features into the nearby 3D space. We resize and concatenate the multi-scale outputs from the 5, 9, 13, 17-th layers as the output feature $\in \mathbb{R}^{{D'\over16} \times {H'\over16} \times {W'\over16} \times 384}$. Since the diffusion of the skeletal feature should not be affected by the human position and orientation in the world coordinate system, we transform the skeletal feature locations to the SMPL coordinate system. Then, the query location $\mathbf{x}$ is also transformed to the SMPL coordinate system, and the corresponding skeletal feature ${s'_{1:C,t}}^{\mathbf{x}}  \in \mathbb{R}^{C\times 384}$ is sampled via trilinear interpolation, and a fully-connected layer reduces the channel-size to 128. The resulting skeletal feature ${s'_{1:C,t}}^{\mathbf{x}}  \in \mathbb{R}^{C\times 128}$  is fed into the following multi-view transformer.

\paragraph{Multi-view transformer.}
The sampled time-augmented skeletal feature  ${s'_{1:C,t}}^{\mathbf{x}}$ is fed into the proposed multi-view transformer to obtain our \textit{meta-time} and \textit{meta-view} representation of the query point $\mathbf{x}$, which is explained in Section \red{3.3} of the manuscript. The overview of the cross-attention between the sampled time-augmented skeletal features and time-specific pixel-aligned features is illustrated in \figref{fig:multiview_transformer}. $d$ is set as 128.

\paragraph{NeRF network.}
\begin{figure}[t]
\centering
\def\arraystretch{0.5}

\includegraphics[width=0.80\linewidth] {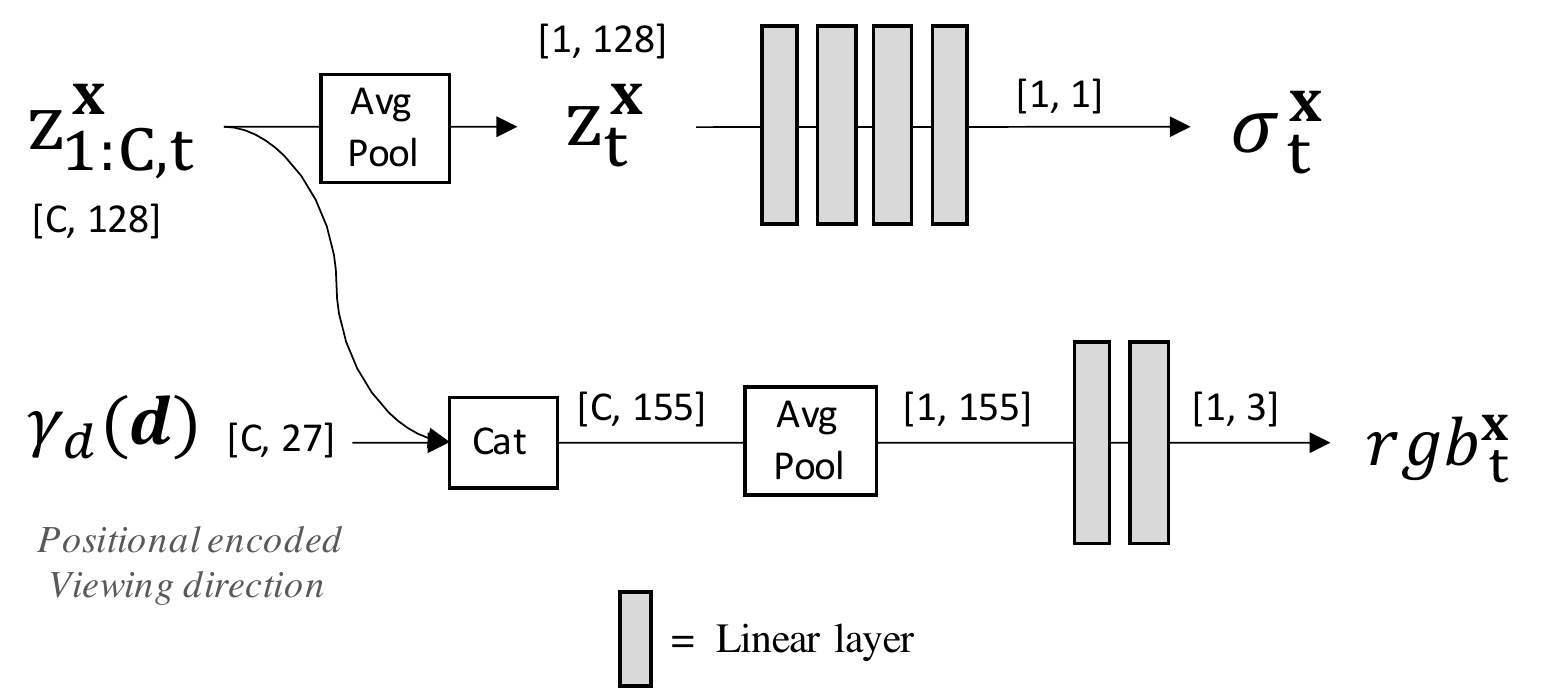}
\caption{\small{\bf Overview of NeRF architecture. } }
\label{fig:nerf_architecture}
\vspace{2mm}
\end{figure}

The NeRF network takes the final representation from above  $z_{1:C,t}^{\mathbf{x}}$ as input and predicts density $\sigma_t^\mathbf{x}$ and color $rgb_t^{\mathbf{x}}$. It consists of the fully-connected layers as illustrated in \figref{fig:nerf_architecture}.

\paragraph{Query point sampling details.}
We first compute the 3D bounding box of the human subject from the corresponding SMPL vertice coordinates. Since there is a gap between the exact human subject geometry and the SMPL model, we enlarge the side length of the bounding box by $2.5\%$ and this becomes the query point sampling bounds. We sample 1024 rays, and 64 points are sampled per ray for the training. For the inference, 64 points are sampled along each ray.

\section{Datasets}
We discuss the additional details about the datasets used, including the train/test splits and license information. Note that both the ZJU-Mocap and AIST datasets do not contain any personally identifiable information or offensive content.

\subsection{ZJU-MoCap}
We use the $512 \times 512$ videos for the training and testing following the original Neural Body \cite{peng2021neural}. ZJU-Mocap provides 10 human subjects, and we reserved 7 for the training and 3 for testing on unseen identities. As mentioned in the main paper, we experiment with 5 independent runs with random train/test splits. For the qualitative results, we used subject 387, 393, 394 for the testing. ZJU-Mocap provides SMPL parameters obtained using EasyMocap\footnote{https://github.com/zju3dv/EasyMocap}~\cite{dong2020motion,peng2021neural,fang2021mirrored,dong2021fast} and foreground mask extracted using PGN \cite{gong2018instance}. ZJU-Mocap is the public dataset that is only meant for the research purposes as stated in their GitHub page.

\subsection{AIST}
The original AIST dataset provides 60 fps videos with $1080 \times 1920$ resolutions \cite{tsuchida2019aist} with corresponding SMPL parameters \cite{li2021learn} obtained using AIST++ API\footnote{https://github.com/google/aistplusplus\_api}. AIST dataset does not provide foreground mask, so we obtained the foreground mask using PGN \cite{gong2018instance}. Since most part of the images are background, we center-crop the video to $800 \times 800$ sizes. During the training and evaluation, we resize the center-cropped video to $512 \times 512$. AIST contains 30 human subjects. We split the train and testing sets based on different subjects, which also makes sure the human motions in the train (20 identities) and testing sets (10 identities) have no overlap. AIST videos are public dataset only for the research purposes. The annotations of the AIST dataset is also public for research purposes and it is licensed by Google LLC CC-BY-4.0 license.

\end{document}